\documentclass[11pt]{article}
\usepackage{acl2016}
\usepackage[utf8]{inputenc} 
\usepackage[T1]{fontenc}    
\usepackage{hyperref}       
\usepackage{url}            
\usepackage{booktabs}       
\usepackage{amsfonts}       
\usepackage{nicefrac}       
\usepackage{microtype}      
\usepackage{times}
\usepackage{url}
\usepackage{latexsym}
\usepackage{amssymb}
\usepackage{lipsum}
\usepackage{mathtools}
\mathtoolsset{showonlyrefs}
\usepackage{graphicx,color}
\usepackage{subcaption}
\usepackage{tikz}
\usepackage{floatrow}
\usepackage{graphicx}
\usepackage{subcaption}
\usepackage{cprotect}
\usepackage[export]{adjustbox}
\usetikzlibrary{shapes,fit,positioning,backgrounds}
\tikzset{>=latex}

\aclfinalcopy

\newfloatcommand{capbtabbox}{table}[][\FBwidth]

\newcommand{\blackcircle}{\raisebox{-.4ex}{\scalebox{1.66}{$\bullet$}}}
\newcommand{\invcircledast}{%
  \mathbin{\vphantom{\circledast}\text{%
    \ooalign{\smash{\blackcircle}\cr
             \hidewidth\smash{\textcolor{white}{$*$}}\hidewidth\cr
            }%
  }}%
}

\title{Neural Associative Memory for Dual-Sequence Modeling}

\author{Dirk Weissenborn\\
	    Language Technology Lab, DFKI\\
	    Alt-Moabit 91c\\
	    Berlin, Germany\\
        {\tt dirk.weissenborn@dfki.de}}

\date{}

\begin{document}
\maketitle
\begin{abstract}
Many important NLP problems can be posed as dual-sequence or sequence-to-sequence modeling tasks. Recent advances in building end-to-end neural architectures have been highly successful in solving such tasks. In this work we propose a new architecture for dual-sequence modeling that is based on associative memory. We derive AM-RNNs, a recurrent associative memory (AM) which augments generic recurrent neural networks (RNN). This architecture is extended to the Dual AM-RNN which operates on two AMs at once. Our models achieve very competitive results on textual entailment. A qualitative analysis demonstrates that long range dependencies between source and target-sequence can be bridged effectively using Dual AM-RNNs. However, an initial experiment on auto-encoding reveals that these benefits are not exploited by the system when learning to solve sequence-to-sequence tasks which indicates that additional supervision or regularization is needed.
\end{abstract}

\section{Introduction}

Dual-sequence modeling and sequence-to-sequence modeling are important paradigms that are used in many applications involving natural language, including machine translation \cite{bahdanau2014neural,sutskever2014sequence}, recognizing textual entailment \cite{cheng2016long,rocktaschel2015reasoning,wang2016learning}, auto-encoding \cite{li2015hierarchical}, syntactical parsing \cite{vinyals2015grammar} or document-level question answering \cite{nips15_hermann}. We might even argue that most, if not all, NLP problems can (at least partially) be modeled by this paradigm \cite{li2015nlp}. These models operate on two distinct sequences, the source and the target sequence. Some tasks require the generation of the target based on the source (sequence-to-sequence modeling), e.g., machine translation, whereas other tasks involve making predictions about a given source and target sequence (dual-sequence modeling), e.g., recognizing textual entailment. Existing state-of-the-art, end-to-end differentiable models for both tasks exploit the same architectural ideas. 

The ability of such models to carry information over long distances is a key enabling factor for their performance. Typically this can be achieved by employing \textit{recurrent neural networks} (RNN) that convey information over time through an internal memory state. Most famous is the LSTM \cite{hochreiter1997long} that accumulates information at every time step additively into its memory state, which avoids the problem of vanishing gradients that hindered previous RNN architectures from learning long range dependencies. For example, \newcite{sutskever2014sequence} connected two LSTMs conditionally for machine translation where the memory state after processing the source was used as initialization for the memory state of the target LSTM. This very simple architecture achieved competitive results compared to existing, very elaborate and feature-rich models. However, learning the inherent long range dependencies between source and target requires extensive training on large datasets. \newcite{bahdanau2014neural} proposed an architecture that resolved this issue by allowing the model to \textit{attend} over all positions in the source sentence when predicting the target sentence, which enabled the model to automatically learn alignments of words and phrases of the source with the target sentence. The important difference is that previous long range dependencies could be bridged directly via attention. However, this architecture requires a larger number of operations that scales with the product of the lengths of the source- and target sequence and a memory that scales with the length of the source sequence.

In this work we introduce a novel architecture for dual-sequence modeling that is based on \textit{associative memories} (AM). AMs are fixed sized memory arrays used to read and write content via an associated keys. Holographic Reduced Representations (HRR) \cite{plate1995holographic}) enable the robust and efficient retrieval of previously written content from redundant memory arrays. Our approach is inspired by the works of \newcite{danihelka2016associative} who recently demonstrated the benefits of exchanging the memory cell of an LSTM with an associative memory on various sequence modeling tasks. In contrast to their architecture which directly adapts the LSTM architecture we propose an augmentation to generic RNNs (\textit{AM-RNNs}, \S\ref{sec:am_rnn}). Similar in spirit to \textit{Neural Turing Machines} \cite{graves2014neural} we decouple the AM from the RNN and restrict the interaction with the AM to read and write operations which we believe to be important. Based on this architecture we derive the \textit{Dual AM-RNN} (\S\ref{sec:dual_assoc_mem}) that operates on two associative memories simultaneously for dual-sequence modeling. We conduct experiments on the task of recognizing textual entailment (\S\ref{sec:experiments}). Our results and qualitative analysis demonstrate that AMs can be used to bridge long range dependencies similar to the attention mechanism while preserving the computational benefits of conveying information through a single, fixed-size memory state. Finally, an initial inspection into sequence-to-sequence modeling with Dual AM-RNNs shows that there are open problems that need to be resolved to make this approach applicable to these kinds of tasks. 

A TensorFlow \cite{tensorflow2015} implementation of (Dual)-AM RNNs can be found at \url{https://github.com/dirkweissenborn/dual_am_rnn}.

\section{Related Work}\label{sec:rel_arch}

Augmenting RNNs by the use of memory is not novel. \newcite{graves2014neural} introduced Neural Turing Machines which augment RNNs with external memory that can be written to and read from. It contains a predefined number of slots to write content to. This form of memory is addressable via content or position shifts. Neural Turing Machines  inspired subsequent work on using different kinds of external memory, like queues or stacks \cite{grefenstette2015learning}. Operations on these memories are calculated via a recurrent controller which is decoupled from the memory whereas AM-RNNs apply the RNN \textit{cell}-function directly upon the content of the associative memory. 

\newcite{danihelka2016associative} introduced Associative LSTMs which extends standard LSTMs directly by reading and writing operations on an associative memory. This architecture is closely related to ours. However, there are crucial differences that are due to the fact that we decouple the associative array from the original \textit{cell}-function. \newcite{danihelka2016associative} directly include operations on the AM in the definition of their Associative LSTM. This might cause problems, since some operations, e.g., \textit{forget}, are directly applied to the entire memory array although this can affect all elements stored in the memory. We believe that only reading and writing operations with respect to a calculated key should be performed on the associative memory. Further operations should therefore only be applied on the stored elements.

Neural attention is another important mechanism that realizes a form of content addressable memory. Most famously it has been applied to machine translation (MT) where attention models automatically learn soft word alignments between source and translation \cite{bahdanau2014neural}. Attention requires memory that stores states of its individual entries, separately, e.g., states for every word in the source sentence of MT or textual entailment \cite{rocktaschel2015reasoning}, or entire sentence states as in \newcite{sukhbaatar2015end} which is an end-to-end memory network \cite{weston2014memory} for question answering. Attention weights are computed based on a provided input and the stored elements. The thereby weighted memory states are summed and the result is retrieved to be used as input to a down-stream neural network. Architectures based on attention require a larger amount of memory and a larger number of operations which scales with the usually dynamically growing memory. In contrast to attention Dual AM-RNNs utilize fixed size memories and a constant number of operations.

AM-RNNs also have an interesting connection to LSTM-Networks \cite{cheng2016long} which recently demonstrated impressive results on various text modeling tasks. LSTM-Networks (LSTMN) select a previous hidden state via attention on a memory tape of past states (intra-attention) opposed to using the hidden state of the previous time step. The same idea is implicitly present in our architecture by retrieving a previous state via a computed key from the associative memory (Equation~\eqref{eq:read}). The main difference lies in the used memory architecture. We use a fixed size memory array in contrast to a dynamically growing memory tape which requires growing computational and memory resources. The drawback of our approach, however, is the potential loss of explicit memories due to retrieval noise or overwriting.

\section{Associative Memory RNN}

\subsection{Redundant Associative Memory}\label{sec:assoc_mem}
In the following, we use the terminology of \newcite{danihelka2016associative} to introduce Redundant Associative Memories and Holographic Reduced Representations (HRR) \cite{plate1995holographic}. HRRs provide a mechanism to encode an item $\boldsymbol{x}$ with a key $\boldsymbol{r}$ that can be written to a fixed size memory array $\boldsymbol{m}$ and that can be retrieved from $\boldsymbol{m}$ via $\boldsymbol{r}$.

In HRR, keys $\boldsymbol{r}$ and values $\boldsymbol{x}$ refer to complex vectors that consist of a \textit{real} and \textit{imaginary} part: $\boldsymbol{r}=\boldsymbol{r}_{re} + i \cdot \boldsymbol{r}_{im}$, $\boldsymbol{x}=\boldsymbol{x}_{re} + i \cdot \boldsymbol{x}_{im}$, where $i$ is the imaginary unit. We represent these complex vectors as concatenations of their respective real and imaginary parts, e.g., $\boldsymbol{r} = [\boldsymbol{r}_{re};\boldsymbol{r}_{im}]$. The encoding- and retrieval-operation proposed by \newcite{plate1995holographic} and utilized by \newcite{danihelka2016associative} is the complex multiplication (Equation~\eqref{eq:complex_mul}) of a key $\boldsymbol{r}$ with its value $\boldsymbol{x}$ (\textit{encoding}), and the complex conjugate of the key $\overline{\boldsymbol{r}}=\boldsymbol{r}_{re} - i \cdot \boldsymbol{r}_{im}$ with the memory (\textit{retrieval}), respectively. Note, that this requires the modulus of the key to be equal to one, i.e., $\sqrt{\boldsymbol{r}_{re} \odot \boldsymbol{r}_{re} + \boldsymbol{r}_{im} \odot \boldsymbol{r}_{im}} = \boldsymbol{1}$, such that $\overline{\boldsymbol{r}} = \boldsymbol{r}^{-1}$. Consider a single memory array $\boldsymbol{m}$ containing $N$ elements $\boldsymbol{x}_k$ with respective keys $\boldsymbol{r}_k$ (Equation~\eqref{eq:memory}).
\begin{align}
    \boldsymbol{r} \circledast \boldsymbol{x} &= 
    \begin{bmatrix} \boldsymbol{r}_{re} \odot \boldsymbol{x}_{re} - \boldsymbol{r}_{im} \odot \boldsymbol{x}_{im} \\
                    \boldsymbol{r}_{re} \odot \boldsymbol{x}_{im} + \boldsymbol{r}_{im} \odot \boldsymbol{x}_{re} \end{bmatrix}  \label{eq:complex_mul} \\
    \boldsymbol{m} &= \sum_{k=1}^N \boldsymbol{r}_k \circledast \boldsymbol{x}_k \label{eq:memory} 
\end{align}

We retrieve an element $\boldsymbol{x}_k$ by multiplying $\overline{\boldsymbol{r}_k}$ with $\boldsymbol{m}$ (Equation~\eqref{eq:retrieval}).
\begin{align}
    \boldsymbol{\tilde{x}}_k &= \overline{\boldsymbol{r}_k} \circledast \boldsymbol{m} = \sum_{k'=1}^N \overline{\boldsymbol{r}_k} \circledast \boldsymbol{r}_{k'} \circledast \boldsymbol{x}_{k'}  \nonumber \\
                &= \boldsymbol{x}_k + \sum_{k'=1\neq k}^N \overline{\boldsymbol{r}_k} \circledast \boldsymbol{r}_{k'} \circledast \boldsymbol{x}_{k'} \nonumber \\
                &= \boldsymbol{x}_k + noise \label{eq:retrieval} 
\end{align}

To reduce noise \newcite{danihelka2016associative} introduce permuted, redundant copies $\boldsymbol{m}_s$ of $ \boldsymbol{m}$ (Equation~\eqref{eq:perm}). This results in uncorrelated retrieval noises which effectively reduces the overall retrieval noise when computing their mean. Consider $N_c$ permutations represented by permutation matrices $P_s$. The retrieval equation becomes the following.
\begin{align}
    \boldsymbol{m}_s &= \sum_{k=1}^N (P_s \boldsymbol{r}_k) \circledast \boldsymbol{x}_k \label{eq:perm} \\
    \boldsymbol{\tilde{x}}_k &= \frac{1}{N_c}\sum_{s=1}^{N_c} \sum_{k'=1}^N (P_s \overline{\boldsymbol{r}_k}) \circledast \boldsymbol{m}_s \nonumber \\
                &= \boldsymbol{x}_k + \sum_{k'=1\neq k}^N \boldsymbol{x}_{k'} \circledast \frac{1}{N_c}   \sum_{s=1}^{N_c} P_s (\overline{\boldsymbol{r}_k} \circledast \boldsymbol{r}_{k'}) \nonumber \\
                &= \boldsymbol{x}_k + noise
\end{align}

The resulting retrieval noise becomes smaller because the mean of the permuted, complex key products tends towards zero with increasing $N_c$ if the key dimensions are uncorrelated (see \newcite{danihelka2016associative} for more information).

\subsection{Augmenting RNNs with Associative Memory}\label{sec:am_rnn}

A recurrent neural network (RNN) can be defined by a parametrized \textit{cell}-function $f_{\boldsymbol{\theta}}: \mathbb{R}^N \times \mathbb{R}^M \to \mathbb{R}^M \times \mathbb{R}^H$ that is recurrently applied to an input sequence $\boldsymbol{X}=(\boldsymbol{x}_1,...,\boldsymbol{x}_T)$. At each time step $t$ it emits an output $\boldsymbol{h}_t$ and a state $\boldsymbol{s}_t$, that is used as additional input in the following time step (Equation~\eqref{eq:rnn}).
\begin{align}
    &f_{\boldsymbol{\theta}}(\boldsymbol{x}_t, \boldsymbol{s}_{t-1}) = (\boldsymbol{s}_t, \boldsymbol{h}_t) \\
    &\boldsymbol{x} \in \mathbb{R}^N,\, \boldsymbol{s}\in\mathbb{R}^M,\, \boldsymbol{h}\in\mathbb{R}^H \label{eq:rnn}
\end{align}

In this work we augment RNNs, or more specifically their \textit{cell}-function $f_{\boldsymbol{\theta}}$, with associative memory to form \textit{Associative Memory RNNs} (AM-RNN) $\tilde{f}_{\boldsymbol{\theta}}$ as follows. Let $\boldsymbol{s}_t = [\boldsymbol{c}_t; \boldsymbol{n}_t]$ be the concatenation of a memory state $\boldsymbol{c}_t$ and, optionally, some remainder $\boldsymbol{n}_t$ that might additionally be used in $f$, e.g., the output of an LSTM. For brevity, we neglect $\boldsymbol{n}_t$ in the following, and thus $\boldsymbol{s}_t = \boldsymbol{c}_t$. At first, we compute a key given the previous output and the current input, which is in turn used to read from the associative memory array $\boldsymbol{m}$ to retrieve a memory state $\boldsymbol{s}$ for the specified key (Equation~\eqref{eq:read}).
\begin{align}
    \boldsymbol{r}_t &= \operatorname{bound} \left( W_r \begin{bmatrix}\boldsymbol{x}_{t}\\ \boldsymbol{h}_{t-1} \end{bmatrix} \right) \nonumber \\
    \boldsymbol{s}_{t-1} &= \overline{\boldsymbol{r}_t} \circledast \boldsymbol{m}_{t-1} \label{eq:read}
\end{align}

The $\operatorname{bound}$-operation \cite{danihelka2016associative} (Equation~\eqref{eq:bound}) guarantees that the modulus of $\boldsymbol{r}_t$ is not greater than $\boldsymbol{1}$. This is an important necessity as mentioned in \S~\ref{sec:assoc_mem}.
\begin{align}
    \operatorname{bound}(\boldsymbol{r}^\prime) &= \begin{bmatrix} \boldsymbol{r}^\prime_{re} \oslash \boldsymbol{d} \\ \boldsymbol{r}^\prime_{im} \oslash \boldsymbol{d} \end{bmatrix}  \label{eq:bound}  \\
    \boldsymbol{d} &= \max\left(1, \sqrt{\boldsymbol{r}^\prime_{re} \odot \boldsymbol{r}^\prime_{re} + \boldsymbol{r}^\prime_{im} \odot \boldsymbol{r}^\prime_{im}} \right)
\end{align}

Next, we apply the original \textit{cell}-function $f_{\boldsymbol{\theta}}$ to the retrieved memory state (Equation~\eqref{eq:apply}) and the concatenation of the current input and last output which serves as input to the internal RNN. We update the associative memory array with the updated state using the conjugate key of the retrieval key (Equation~\eqref{eq:update}).
\begin{align}
    \boldsymbol{s}_t, \boldsymbol{h}_t &= f_{\boldsymbol{\theta}}\left(\begin{bmatrix}\boldsymbol{x}_{t}\\ \boldsymbol{h}_{t-1} \end{bmatrix}, \boldsymbol{s}_{t-1} \right) \label{eq:apply} \\
    \boldsymbol{m}_{t} &= \boldsymbol{m}_{t-1} + \boldsymbol{r}_t \circledast (\boldsymbol{s}_t - \boldsymbol{s}_{t-1}) \nonumber \\
    \tilde{f}_{\boldsymbol{\theta}}(\boldsymbol{x}_t, \boldsymbol{m}_{t-1}) &= (\boldsymbol{m}_t, \boldsymbol{h}_t) \label{eq:update}
\end{align}

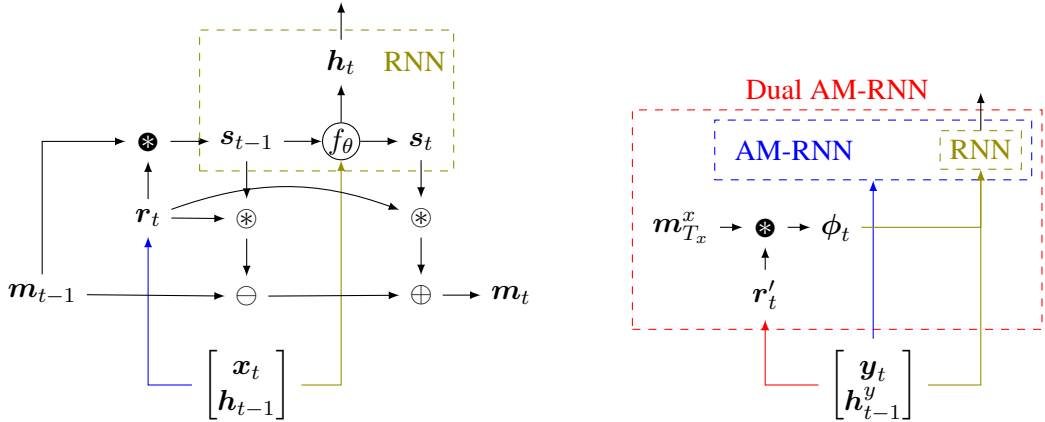
\begin{figure*}[t]
\begin{subfigure}[t]{0.48\textwidth}
\centering
\begin{tikzpicture}[->]
\tikzstyle{operator}=[circle, black, thick, inner sep=0pt]

\node (AM1) {$\boldsymbol{m}_{t-1}$};

\node[above right=5mm and 5mm of AM1] (r) {$\boldsymbol{r}_{t}$};

\node[above right=5mm and 5mm of r] (c1) {$\boldsymbol{s}_{t-1}$};
\node[right=5mm of c1, draw, circle, inner sep=0pt] (f) {$f_{\theta}$};
\node[right=5mm of f] (c2) {$\boldsymbol{s}_{t}$};
\node[above=5mm of f] (h) {$\boldsymbol{h}_{t}$};
\node[above=5mm of h] (out) {};
\node[above=5mm of r] (read) {$\invcircledast$};
\node[below=5mm of c1] (write1) {$\circledast$};
\node[below=5mm of c2] (write2) {$\circledast$};
\node[below=5mm of write2] (add) {$\oplus$};
\node[below=5mm of write1] (sub) {$\ominus$};

\node[below=3mm of sub] (x) {$\begin{bmatrix}\boldsymbol{x}_{t}\\ \boldsymbol{h}_{t-1} \end{bmatrix}$};

\node[right=5mm of add] (AM2) {$\boldsymbol{m}_{t}$};

\path [draw] (AM1) |- (read);
\path (r) edge[] (read);
\path (read) edge[] (c1);

\path [draw,color=blue] (x) -| (r);

\path (c1) edge[] (f);
\path [draw, color=olive] (x) -| (f);
\path (f) edge[] (c2);
\path (f) edge[] (h);
\path (AM1) edge[] (sub);
\path (r) edge[bend left=20] (write2);
\path (c2) edge[] (write2);
\path (r) edge[] (write1);
\path (c1) edge[] (write1);
\path (write1) edge[] (sub);
\path (write2) edge[] (add);
\path (sub) edge[] (add);
\path (add) edge[] (AM2);
\path (h) edge[] (out);

\node[draw,dashed,fit=(c1) (c2) (h),color=olive] (rnn) {};
\node[right=1mm of h] {\textcolor{olive}{RNN}};
\end{tikzpicture}
\caption{Illustration of AM-RNN for input $\boldsymbol{x}_t$ at time step $t$.}\label{fig:AMRNN}
\end{subfigure}~
\begin{subfigure}[t]{0.43\textwidth}
\centering
\begin{tikzpicture}[->]
\tikzstyle{operator}=[circle, black, thick, inner sep=0pt]

\node (alpha) {$\boldsymbol{m}^x_{T_x}$};

\node[right=3mm of alpha] (read) {$\invcircledast$};

\node[below=3mm of read] (r) {$\boldsymbol{r}^\prime_t$};
\node[right=3mm of read] (phi) {$\boldsymbol{\phi}_{t}$};
\node[above right=5mm and 2cm of read, draw, dashed, color=olive] (rnn) {RNN};
\node[left=10mm of rnn] (AM_label) {\textcolor{blue}{AM-RNN}};
\node[fit=(AM_label) (rnn), draw, dashed, color=blue] (AM) {};

\node[above=5mm of rnn](out) {};

\node[draw,dashed,fit=(alpha) (AM) (r) (AM_label), color=red] (DAM) {};
\node[above=0mm of DAM] {\textcolor{red}{Dual AM-RNN}};

\node[below=21mm of AM] (y) {$\begin{bmatrix}\boldsymbol{y}_{t}\\ \boldsymbol{h}^y_{t-1} \end{bmatrix}$};

\path [draw] (alpha) |- (read);
\path (r) edge (read);
\path (read) edge (phi);
\path (rnn) edge (out);
\path (y) edge[color=blue] (AM);
\path [draw, color=olive] (y) -| (rnn);
\path [draw, color=olive] (phi) -| (rnn);
\path [draw, color=red] (y) -| (r);
\end{tikzpicture}
\caption{Illustration of a Dual AM-RNN that extends the AM-RNN with the utilization of the final memory array $\boldsymbol{m}^x_{T_x}$ of source sequence $\boldsymbol{X}$. }\label{fig:DAMRNN}
\end{subfigure}
\cprotect\caption{Illustration of the computation workflow in AM-RNNs and Dual AM-RNNs.
$\invcircledast$ refers to the complex multiplication with the (complex) conjugate of $\boldsymbol{r}^\cdot_t$ and can be interpreted as the retrieval operation. Similarly, $\circledast$ can be interpreted as the encoding operation.}
\end{figure*}

The entire computation workflow is illustrated in Figure~\ref{fig:AMRNN}.

\section{Associative Memory RNNs for Dual Sequence Modeling}\label{sec:dual_assoc_mem}

Important NLP tasks such as machine translation (MT) or detecting textual entailment (TE) involve two distinct sequences as input, a source- and a target sequence. In MT a system predicts the target sequence based on the source whereas in TE source and target are given and an entailment-class should be predicted. Recently, both tasks were successfully modelled using an attention mechanism that can attend over positions in the source sentence at any time step in the target sentence \cite{bahdanau2014neural,rocktaschel2015reasoning,cheng2016long}. These models are able to learn important task specific correlations between words or phrases of the two sentences, like word/phrase translation, or word-/phrase-level entailment or contradiction. The success of these models is mainly due to the fact that long range dependencies can be bridged directly via attention, instead of keeping information over long distances in a memory state that can get overwritten. 

The same can be achieved through associative memory. Given the correct key a state that was written at any time step in the source sentence can be retrieved from an AM with minor noise that can efficiently be reduced by redundancy. Therefore, AMs can bridge long range dependencies and can therefore be used as an alternative to attention. The trade-off for using an AM is that memorized states cannot be used for their retrieval. However, the retrieval operation is constant in time and memory whereas the computational and memory complexity of attention based architectures grow linearly with the length of the source sequence.

We propose two different architectures for solving dual sequence problems. Both approaches use at least one AM-RNN for processing the source and another for the target sequence. The first approach reads the source sequence $\boldsymbol{X}=(\boldsymbol{x}_1,...,\boldsymbol{x}_{T_x})$ and uses the final associative memory array $\boldsymbol{m}^x(:=\boldsymbol{m}^x_{T_x})$ to initialize the memory array $\boldsymbol{m}^y_0=\boldsymbol{m}^x$ of the AM-RNN that processes the target sequence $\boldsymbol{Y}=(\boldsymbol{y}_1,...,\boldsymbol{y}_{T_y})$. Note that this is basically the the conditional encoding architecture of \newcite{rocktaschel2015reasoning}. 

The second approach uses the final AM array of the source sequence $\boldsymbol{m}^x$ in addition to an independent target AM array $\boldsymbol{m}^y_t$. At each time step $t$ the \textit{Dual AM-RNN} computes another key $\boldsymbol{r}^\prime_{t}$ that is used to read from $\boldsymbol{m}^x$ and feeds the retrieved value as additional input to $\boldsymbol{y}_t$ to the inner RNN of the target AM-RNN. These changes are reflected in the Equation~\eqref{eq:apply_dual} (compared to Equation~\eqref{eq:apply}) and illustrated in Figure~\ref{fig:DAMRNN}.
\begin{align}
    \boldsymbol{r}^\prime_t &= \operatorname{bound} \left( W_{r^\prime} \begin{bmatrix}\boldsymbol{y}_{t}\\ \boldsymbol{h}^y_{t-1} \end{bmatrix} \right) \nonumber \\
    \boldsymbol{\phi}_{t} &= \overline{\boldsymbol{r}^\prime_t} \circledast \boldsymbol{m}^x \nonumber \\
    \boldsymbol{s}_t, \boldsymbol{h}^y_t &= f_{\boldsymbol{\theta}}\left(\begin{bmatrix}\boldsymbol{y}_{t}\\ \boldsymbol{h}^y_{t-1} \\ \boldsymbol{\phi}_{t} \end{bmatrix}, \boldsymbol{s}_{t-1} \right) \label{eq:apply_dual} \\
\end{align}

\section{Experiments}\label{sec:experiments}

\subsection{Setup}

\paragraph{Dataset} We conducted experiments on the Stanford Natural Language Inference (SNLI) Corpus \cite{snli:emnlp2015} that consists of roughly $500$k sentence pairs (premise-hypothesis). They are annotated with textual entailment labels. The task is to predict whether a premise \textit{entails}, \textit{contradicts} or is \textit{neutral} to a given hypothesis.

\paragraph{Training} 
We perform mini-batch ($B=50$) stochastic gradient descent using ADAM \cite{kingma2014adam} with $\beta_1=0$, $\beta_2=0.999$ and an initial learning rate of $10^{-3}$ for small models ($H \approx 100$) and $10^{-4}$ ($H=500$) for our large model. The learning rate was halved whenever accuracy dropped over the period of one epoch. Performance on the development set was checked every 1000 mini-batches and the best model is used for testing. We employ dropout with a probability of $0.1$ or $0.2$ for the small and large models, respectively.
Following \newcite{cheng2016long}, word embeddings are initialized with \verb|Glove| \cite{pennington2014glove} or randomly for unknown words. \verb|Glove| initialized embeddings are tuned only after an initial epoch through the training set. 

\paragraph{Model} In this experiment we compare the traditional GRU with the (Dual) AM-GRU using conditional encoding \cite{rocktaschel2015reasoning} using shared parameters between source and target RNNs. Associative memory is implemented with 8 redundant memory copies. For the Dual AM-GRU we define $\boldsymbol{r}^\prime_t = \boldsymbol{r}_t$ (see \S~\ref{sec:dual_assoc_mem}), i.e., we use the same key for interacting with the premise and hypothesis associative memory array while processing the hypothesis. The rationale behind this is that we want to retrieve text passages from the premise that are similar to text passages of the target sequence.

All of our models consist of 2 layers with a GRU as top-layer which is intended to summarize outputs of the bottom layer. The bottom layer corresponds to our different architectures. We concatenate the final output of the premise and hypothesis together with their absolute difference to form the final representation that is used as input to a two-layer perceptron with rectifier-activations for classification.

\subsection{Results}

\begin{table*}[t]
\begin{tabular}{l c c c}
   \toprule
   \textbf{Model} & $H / |\boldsymbol{\theta}_{-E}|$ & \textbf{Accuracy} \\ 
   \midrule
   LSTM \cite{rocktaschel2015reasoning} & $116/252$k & $80.9$ \\
   LSTM shared \cite{rocktaschel2015reasoning} & $159/252$k & $81.4$ \\
   LSTM-Attention \cite{rocktaschel2015reasoning} & $100/252$k & $\mathbf{83.5}$ \\
   \midrule
   GRU shared & $126/321$k & $81.9$ \\
   AM-GRU shared & $108/329$k & $82.9$ \\
   Dual AM-GRU shared & $100/321$k & $\mathbf{84.4}$ \\
   \midrule
   Dual AM-GRU shared & $500/5.6$m & $\underline{85.4}$ \\
   LSTM Network \cite{cheng2016long} & $450/3.4$m & $\mathbf{86.3}$ \\
   \bottomrule
\end{tabular}
\caption{Accuracies of different RNN-based architectures on SNLI dataset. We also report the respective hidden dimension $H$ and number of parameters $|\boldsymbol{\theta}_{-E}|$ for each architecture without taking word embeddings $E$ into account.}\label{tab:results_snli}
\end{table*}

The results are presented in Table~\ref{tab:results_snli}. They long range that the $H$=100-dimensional Dual AM-GRU and conditional AM-GRU outperform our baseline GRU system significantly. Especially the Dual AM-GRU does very well on this task achieving $84.4\%$ accuracy, which shows that it is important to utilize the associative memory of the premise separately for reading only. Most notably is that it achieves even better results than a comparable LSTM architecture with two-way attention between all premise and hypothesis words (LSTM-Attention). This indicates that our Dual AM-GRU architecture is at least able to perform similar or even better than an attention-based model in this setup.

\begin{figure*}[p]
    \centering
    \begin{subfigure}[l]{0.5\textwidth}
        \includegraphics[width=0.9\textwidth,frame]{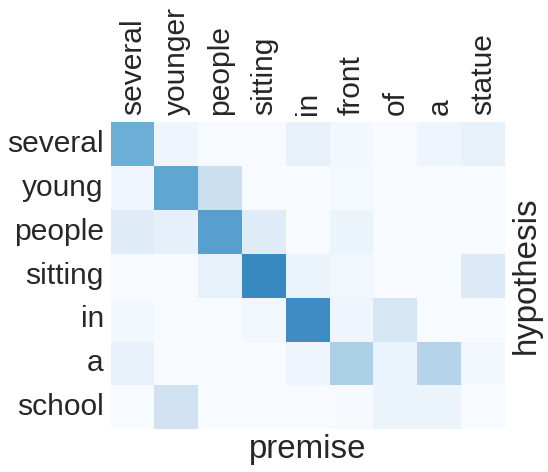}
    \end{subfigure}~
    \begin{subfigure}[r]{0.5\textwidth}
        \includegraphics[width=0.9\textwidth,frame]{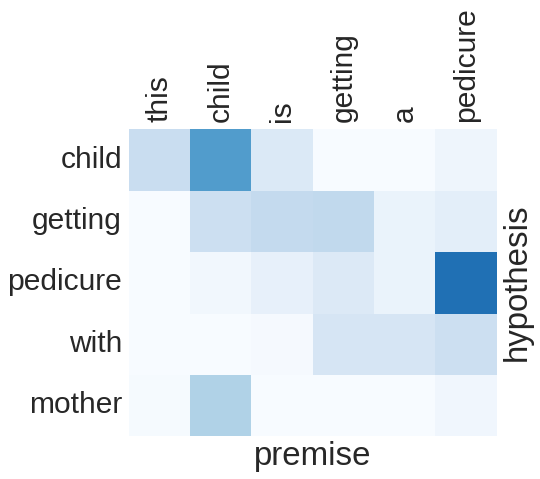}
    \end{subfigure}
    
    \hfill
    
    \begin{subfigure}[l]{0.5\textwidth}
        \includegraphics[width=0.9\textwidth,frame]{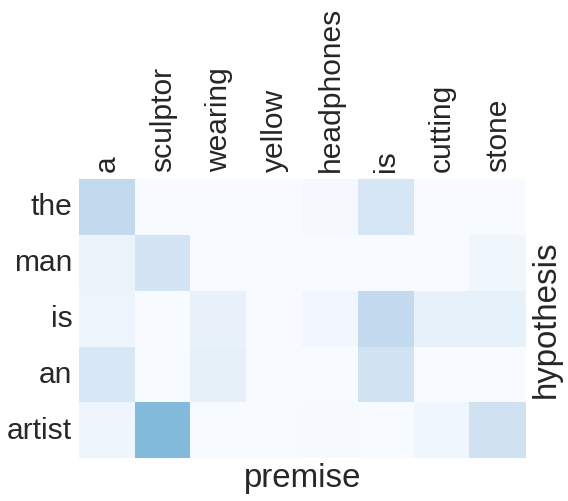}
    \end{subfigure}~
    \begin{subfigure}[r]{0.5\textwidth}
        \includegraphics[width=0.9\textwidth,frame]{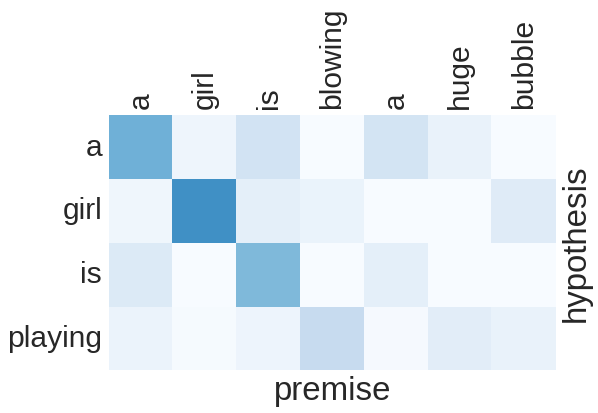}
    \end{subfigure}
    
    \hfill
    
    \begin{subfigure}[l]{0.5\textwidth}
    \centering
         \includegraphics[width=0.7\textwidth,frame]{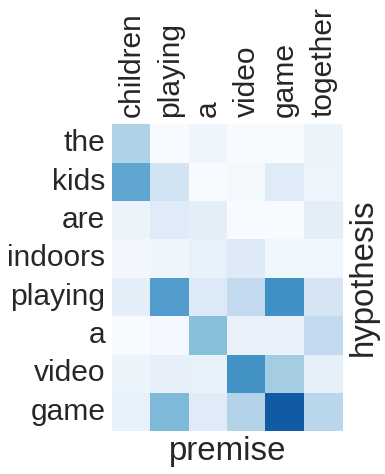}
    \end{subfigure}~
    \begin{subfigure}[r]{0.5\textwidth}
        \includegraphics[width=0.9\textwidth,frame]{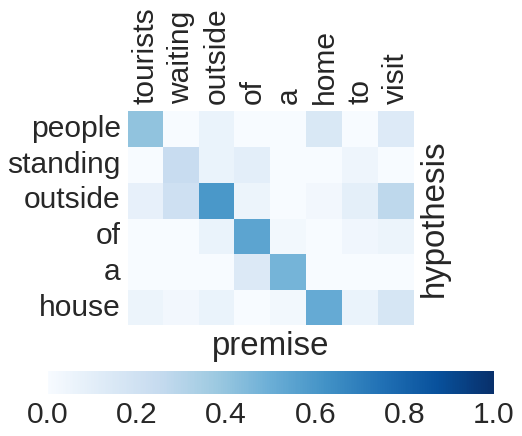}
    \end{subfigure}
    \hfill
    \caption{Heatmaps of cosine similarity between content that has been written to the associative memory at each time step of the premise (x-axis) and what has been retrieved from it by the Dual AM-GRU while processing the hypothesis (y-axis). } \label{fig:illustration}%
\end{figure*}

We investigated this finding qualitatively from sampled examples by plotting heatmaps of cosine similarities between the content that has been written to memory at every time step in the premise and what has been retrieved from it while the Dual AM-GRU processes the hypothesis. Random examples are shown in Figure~\ref{fig:illustration}, where we can see that the Dual AM-GRU is indeed able to retrieve the content from the premise memory that is most related with the respective hypothesis words, thus allowing to bridge important long-range dependencies for solving this task similar to attention. We observe that content for related words and phrases is retrieved from the premise memory when processing the hypothesis, e.g., ``play" and ``video game" or ``artist" and ``sculptor".

Increasing the size of the hidden dimension to 500 improves accuracy by another percentage point. 
The recently proposed LSTM Network achieves slightly better results. However, its number of operations scales with the square of the summed source and target sequence, which is even larger than traditional attention.

\subsection{Sequence-to-Sequence Modeling}

End-to-end differentiable sequence-to-sequence models consist of an encoder that encodes the source sequence and a decoder which produces the target sequence based on the encoded source. In a preliminary experiment we applied the Dual AM-GRU without shared parameters to the task of auto-encoding where source- and target sequence are the same. Intuitively we would like the AM-GRU to write phrase-level information with different keys to the associative memory.
However, we found that the encoder AM-GRU learned very quickly to write everything with the same key to memory, which makes it work very similar to a standard RNN based encoder-decoder architecture where the encoder state is simply used to initialize the decoder state.

\begin{figure}[t]
\centering
\small
\includegraphics[width=\textwidth]{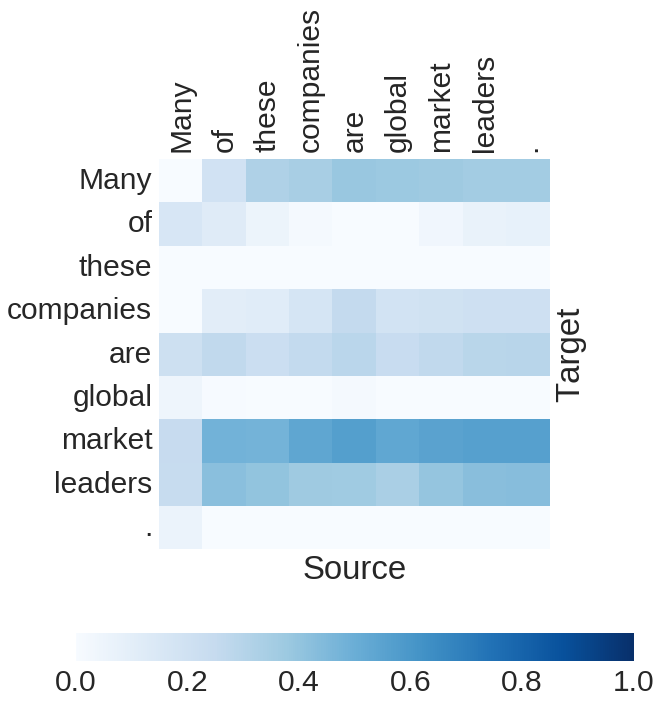}
\caption{Heatmap of cosine similarity between content that has been written to the associative memory at each time step by the encoder (x-axis) and what has been retrieved from it by the Dual AM-GRU while decoding (y-axis).}\label{fig:auto_encode}%
\end{figure}

This finding is illustrated in Figure~\ref{fig:auto_encode}. The presented heatmap shows similarities between content that has been retrieved while predicting the target sequence and what has been written by the encoder to memory. We observe that the similarities between retrieved content and written content are horizontally slightly increasing, i.e., towards the end of the encoded source sentence. This indicates that the encoder overwrites the the associative memory while processing the source with the same key.

\subsection{Discussion}

Our experiments on entailment show that the idea of using associative memory to bridge long term dependencies for dual-sequence modeling can work very well. However, this architecture is not naively transferable to the task of sequence-to-sequence modeling. We believe that the main difficulty lies in the computation of an appropriate key at every time step in the target sequence to retrieve related content. Furthermore, the encoder should be enforced to not always use the same key. For example, keys could be based on syntactical and semantical cues, which might ultimately result in capturing some form of Frame Semantics \cite{fillmore2001frame}. This could facilitate decoding significantly. We believe that this might be achieved via regularization or by curriculum learning \cite{bengio2009curriculum}. 

\section{Conclusion}
We introduced the Dual AM-RNN, a recurrent neural architecture that operates on associative memories. The AM-RNN augments traditional RNNs generically with associative memory. The Dual AM-RNN extends AM-RNNs with a second read-only memory. Its ability to capture long range dependencies enables effective learning of dual-sequence modeling tasks such as recognizing textual entailment. Our models achieve very competitive results and outperform a comparable attention-based model while preserving constant computational and memory resources. Applying the Dual AM-RNN to a sequence-to-sequence modeling task revealed that the benefits of bridging long range dependencies cannot yet be achieved for this kind of problem. However, quantitative as well as qualitative results on textual entailment are very promising and therefore we believe that the Dual AM-RNN can be an important building block for NLP tasks involving two sequences.

\section*{Acknowledgments}
We thank Sebastian Krause, Tim Rockt\"aschel and Leonhard Hennig for comments on an early draft of this work. This research was supported by the German Federal Ministry of Education and 
Research (BMBF) through the projects ALL SIDES (01IW14002), BBDC (01IS14013E), and Software Campus (01IS12050, sub-project GeNIE).

\bibliography{acl2016}

\begin{thebibliography}{}

\bibitem[\protect\citename{Abadi \bgroup et al.\egroup }2015]{tensorflow2015}
Mart\'{\i}n Abadi, Ashish Agarwal, Paul Barham, Eugene Brevdo, Zhifeng Chen,
  Craig Citro, Greg~S. Corrado, Andy Davis, Jeffrey Dean, Matthieu Devin,
  Sanjay Ghemawat, Ian Goodfellow, Andrew Harp, Geoffrey Irving, Michael Isard,
  Yangqing Jia, Rafal Jozefowicz, Lukasz Kaiser, Manjunath Kudlur, Josh
  Levenberg, Dan Man\'{e}, Rajat Monga, Sherry Moore, Derek Murray, Chris Olah,
  Mike Schuster, Jonathon Shlens, Benoit Steiner, Ilya Sutskever, Kunal Talwar,
  Paul Tucker, Vincent Vanhoucke, Vijay Vasudevan, Fernanda Vi\'{e}gas, Oriol
  Vinyals, Pete Warden, Martin Wattenberg, Martin Wicke, Yuan Yu, and Xiaoqiang
  Zheng.
\newblock 2015.
\newblock {TensorFlow}: Large-scale machine learning on heterogeneous systems.
\newblock Software available from tensorflow.org.

\bibitem[\protect\citename{Bahdanau \bgroup et al.\egroup
  }2015]{bahdanau2014neural}
Dzmitry Bahdanau, Kyunghyun Cho, and Yoshua Bengio.
\newblock 2015.
\newblock Neural machine translation by jointly learning to align and
  translate.
\newblock In {\em The International Conference on Learning Representations
  (ICLR)}.

\bibitem[\protect\citename{Bengio \bgroup et al.\egroup
  }2009]{bengio2009curriculum}
Yoshua Bengio, J{\'e}r{\^o}me Louradour, Ronan Collobert, and Jason Weston.
\newblock 2009.
\newblock Curriculum learning.
\newblock In {\em Proceedings of the 26th annual international conference on
  machine learning}, pages 41--48. ACM.

\bibitem[\protect\citename{Bowman \bgroup et al.\egroup }2015]{snli:emnlp2015}
Samuel~R. Bowman, Gabor Angeli, Christopher Potts, and Christopher~D. Manning.
\newblock 2015.
\newblock A large annotated corpus for learning natural language inference.
\newblock In {\em Proceedings of the 2015 Conference on Empirical Methods in
  Natural Language Processing (EMNLP)}. Association for Computational
  Linguistics.

\bibitem[\protect\citename{Cheng \bgroup et al.\egroup }2016]{cheng2016long}
Jianpeng Cheng, Li~Dong, and Mirella Lapata.
\newblock 2016.
\newblock Long short-term memory-networks for machine reading.
\newblock {\em arXiv preprint arXiv:1601.06733}.

\bibitem[\protect\citename{Danihelka \bgroup et al.\egroup
  }2016]{danihelka2016associative}
Ivo Danihelka, Greg Wayne, Benigno Uria, Nal Kalchbrenner, and Alex Graves.
\newblock 2016.
\newblock Associative long short-term memory.
\newblock {\em arXiv preprint arXiv:1602.03032}.

\bibitem[\protect\citename{Fillmore and Baker}2001]{fillmore2001frame}
Charles~J Fillmore and Collin~F Baker.
\newblock 2001.
\newblock Frame semantics for text understanding.
\newblock In {\em Proceedings of WordNet and Other Lexical Resources Workshop,
  NAACL}.

\bibitem[\protect\citename{Graves \bgroup et al.\egroup
  }2014]{graves2014neural}
Alex Graves, Greg Wayne, and Ivo Danihelka.
\newblock 2014.
\newblock Neural turing machines.
\newblock {\em arXiv preprint arXiv:1410.5401}.

\bibitem[\protect\citename{Grefenstette \bgroup et al.\egroup
  }2015]{grefenstette2015learning}
Edward Grefenstette, Karl~Moritz Hermann, Mustafa Suleyman, and Phil Blunsom.
\newblock 2015.
\newblock Learning to transduce with unbounded memory.
\newblock In {\em Advances in Neural Information Processing Systems}, pages
  1819--1827.

\bibitem[\protect\citename{Hermann \bgroup et al.\egroup }2015]{nips15_hermann}
Karl~Moritz Hermann, Tom\'a\v{s} Ko\v{c}isk\'y, Edward Grefenstette, Lasse
  Espeholt, Will Kay, Mustafa Suleyman, and Phil Blunsom.
\newblock 2015.
\newblock Teaching machines to read and comprehend.
\newblock In {\em Advances in Neural Information Processing Systems (NIPS)}.

\bibitem[\protect\citename{Hochreiter and Schmidhuber}1997]{hochreiter1997long}
Sepp Hochreiter and J{\"u}rgen Schmidhuber.
\newblock 1997.
\newblock Long short-term memory.
\newblock {\em Neural computation}, 9(8):1735--1780.

\bibitem[\protect\citename{Kingma and Ba}2015]{kingma2014adam}
Diederik Kingma and Jimmy Ba.
\newblock 2015.
\newblock Adam: A method for stochastic optimization.
\newblock In {\em The International Conference on Learning Representations
  (ICLR)}.

\bibitem[\protect\citename{Li and Hovy}2015]{li2015nlp}
Jiwei Li and Eduard Hovy.
\newblock 2015.
\newblock The {NLP} engine: A universal turing machine for nlp.
\newblock {\em arXiv preprint arXiv:1503.00168}.

\bibitem[\protect\citename{Li \bgroup et al.\egroup }2015]{li2015hierarchical}
Jiwei Li, Minh-Thang Luong, and Dan Jurafsky.
\newblock 2015.
\newblock A hierarchical neural autoencoder for paragraphs and documents.
\newblock In {\em 53nd Annual Meeting of the Association for Computational
  Linguistics (ACL)}.

\bibitem[\protect\citename{Pennington \bgroup et al.\egroup
  }2014]{pennington2014glove}
Jeffrey Pennington, Richard Socher, and Christopher~D Manning.
\newblock 2014.
\newblock Glove: Global vectors for word representation.
\newblock In {\em EMNLP}, volume~14, pages 1532--1543.

\bibitem[\protect\citename{Plate}1995]{plate1995holographic}
Tony~A Plate.
\newblock 1995.
\newblock Holographic reduced representations.
\newblock {\em Neural networks, IEEE transactions on}, 6(3):623--641.

\bibitem[\protect\citename{Rockt{\"a}schel \bgroup et al.\egroup
  }2016]{rocktaschel2015reasoning}
Tim Rockt{\"a}schel, Edward Grefenstette, Karl~Moritz Hermann, Tom{\'a}{\v{s}}
  Ko{\v{c}}isk{\`y}, and Phil Blunsom.
\newblock 2016.
\newblock Reasoning about entailment with neural attention.
\newblock {\em The International Conference on Learning Representations
  (ICLR)}.

\bibitem[\protect\citename{Sukhbaatar \bgroup et al.\egroup
  }2015]{sukhbaatar2015end}
Sainbayar Sukhbaatar, Jason Weston, Rob Fergus, et~al.
\newblock 2015.
\newblock End-to-end memory networks.
\newblock In {\em Advances in Neural Information Processing Systems}, pages
  2431--2439.

\bibitem[\protect\citename{Sutskever \bgroup et al.\egroup
  }2014]{sutskever2014sequence}
Ilya Sutskever, Oriol Vinyals, and Quoc~V Le.
\newblock 2014.
\newblock Sequence to sequence learning with neural networks.
\newblock In {\em Advances in neural information processing systems}, pages
  3104--3112.

\bibitem[\protect\citename{Vinyals \bgroup et al.\egroup
  }2015]{vinyals2015grammar}
Oriol Vinyals, {\L}ukasz Kaiser, Terry Koo, Slav Petrov, Ilya Sutskever, and
  Geoffrey Hinton.
\newblock 2015.
\newblock Grammar as a foreign language.
\newblock In {\em Advances in Neural Information Processing Systems}, pages
  2755--2763.

\bibitem[\protect\citename{Wang and Jiang}2016]{wang2016learning}
Shuohang Wang and Jing Jiang.
\newblock 2016.
\newblock Learning natural language inference with lstm.
\newblock In {\em Proceedings of the 2016 Human Language Technology Conference
  of the North American Chapter of the Association of Computational
  Linguistics}.

\bibitem[\protect\citename{Weston \bgroup et al.\egroup
  }2015]{weston2014memory}
Jason Weston, Sumit Chopra, and Antoine Bordes.
\newblock 2015.
\newblock Memory networks.
\newblock In {\em The International Conference on Learning Representations
  (ICLR)}.

\end{thebibliography}
\bibliographystyle{acl2016}

\end{document}